\documentclass[journal,twoside,web]{ieeecolor}
\usepackage{jsen}
\usepackage{cite}
\usepackage{epsfig} 
\usepackage{times}
\usepackage{amsmath,amssymb,amsfonts} 
\usepackage{algorithmic}
\usepackage{graphicx}
\usepackage{textcomp}
\usepackage{wrapfig}
\usepackage{color}
\usepackage{bm}
\usepackage{multirow}
\usepackage{booktabs}
\usepackage{array}
\usepackage{gensymb}
\usepackage{tabularx}
\usepackage{float}
\usepackage{microtype}
\def\BibTeX{{\rm B\kern-.05em{\sc i\kern-.025em b}\kern-.08em    T\kern-.1667em\lower.7ex\hbox{E}\kern-.125emX}}
\definecolor{abstractbg}{rgb}{0.89804,0.94510,0.83137}
\title{\LARGE \bf
Spatio-temporal encoding  improves neuromorphic \\ tactile texture classification}
\author{Anupam K. Gupta$^{1,\star, \dagger}$, Andrei Nakagawa$^{2}$, Nathan F. Lepora$^{3, 4, \ddagger}$ and Nitish V. Thakor$^{1,5, \ddagger}$
\thanks{This project was supported by grants from Office of Naval Research Global (R719000007597), National Science Foundation (18494117) and National University of Singapore.}
\thanks{$^{1}$ The N.1 Institute of Health, National University of Singapore, Singapore 117456, Singapore
        {\tt\small}}%
\thanks{$^{2}$ Faculty of Electrical Engineering, Federal University of Uberlandia, Uberlandia 38400-902, Brazil
{\tt\small}}%
\thanks{$^{3}$ Department of Engineering Mathematics, University of Bristol, United Kingdom
{\tt\small}}%
\thanks{$^{4}$ Bristol Robotics Laboratory, University of Bristol, United Kingdom
{\tt\small }}%
\thanks{$^{5}$ Department of Biomedical Engineering, The Johns Hopkins University, Baltimore, MD 21218, USA
        {\tt\small}}%
\thanks{$^{\star}$ Corresponding author: anupam.gupta@bristol.ac.uk
{\tt\small}}%
\thanks{$^{\dagger}$ This author has relocated to Bristol Robotics Laboratory, University of Bristol, United Kingdom
{\tt\small}}%
\thanks{$^{\ddagger}$ Joint senior authors}
}

\begin{document}
\IEEEtitleabstractindextext{%
\fcolorbox{abstractbg}{abstractbg}{%
\begin{minipage}{\textwidth}%
\begin{abstract}
With the increase in interest in the deployment of robots in unstructured environments to work alongside humans, the development of human-like sense of touch for robots becomes important. In this work, we implement a multi-channel neuromorphic tactile system that encodes contact episodes as discrete spike events that mimic the behavior of slow adapting mechanoreceptors. We study the impact of information pooling across artificial mechanoreceptors on classification performance of spatially non-uniform naturalistic textures. We encoded the spatio-temporal activation patterns of mechanoreceptors through gray-level co-occurrence matrix computed from time-varying mean spiking rate-based tactile response volume. We found that this approach greatly improved texture classification in comparison to the use of individual mechanoreceptor response alone. In addition, the performance was also more robust to changes in sliding velocity. The importance of exploiting precise spatial and temporal correlations between sensory channels is evident from the fact that on either removal of precise temporal information or altering of spatial structure of response pattern, a significant performance drop was observed. This study thus demonstrates the superiority of population coding approaches that can exploit the precise spatio-temporal information encoded in activation patterns of mechanoreceptor populations. It, therefore, makes an advance in the direction of development of bio-inspired tactile systems required for realistic touch applications in robotics and prostheses.
\end{abstract}

\end{minipage}}}

\maketitle

\section{Introduction}

\subsection{Motivation}

The sense of touch is one of the most primal senses to develop in humans and is behind our remarkable ability to interact, explore and manipulate the environment at will. Absence of touch, will impair us in successfully executing even basic activities necessary for survival. The sense of touch combined with proprioception also contributes immensely in early stages of human development and learning. It is shown to be necessary for early development of shape bias in children necessary for quick learning of novel object categories~\cite{c1}. It assumes even more importance in scenarios where vision is absent or impaired due to low illumination, occlusion etc. For example, imagine navigating around a dark room without the sense of touch.  

Robots, however, are far from acquiring human-like touch capabilities. With an increase in interest in the deployment of robots in unstructured environments, to work alongside humans (for e.g. as nursing / surgical robots) or to restore touch in human amputees~\cite{c2}, it is imperative to develop artificial touch systems that can match their human counterparts. The past few years have seen rapid development of touch sensors that are: i) compliant like human skin~\cite{c3,c4}, ii) mimic biological fingertip receptor density~\cite{c5}, iii) encapsulated in soft sensor coverings imprinted with skin like micro-structures~\cite{c6,c7} and iv) neuromorphic that encode artificial sensor outputs as discrete spike events to mimic mechanoreceptor outputs~\cite{c8,c9,c10,c11}. Bridging the gap between biological touch and artificial touch will require the development of biomimetic systems that includes sensors that resemble their biological counterparts in form and function as well as brain inspired learning algorithms and hardware for efficient and fast learning.

In this work, we focus on natural texture classification, an important surface property estimated through touch and important for myriad of tactile tasks like object recognition. The studies of human touch have shown texture perception limit in humans to be as low as 10 nm~\cite{c12}. Furthermore, the perception of texture, in humans, is found to be invariant to sliding speed~\cite{c13}. The mechanisms that mediate such speed-invariant fine texture perception are not fully known. Due to spatial and temporal capacity constraints, it is not feasible to distinguish textures based on the encoding of absolute vibration frequencies elicited by the textures. Apart from capacity constraints, direct encoding of absolute frequencies is not feasible for two reasons: 1) the mechanism to distinguish different vibratory frequencies is crude with discrimination thresholds of only about 20\%~\cite{c14,c15} and 2) to distinguish textures based on encoding of absolute frequencies, speed-invariant perception would require availability of sliding speed, to compensate for distortion in spike trains, due to variations in sliding speed~\cite{c13}. It has been previously argued, that instead of directly encoding vibratory frequencies, texture perception may rely on extracting abstract structure in the pattern of activation of mechanoreceptor populations instead~\cite{c16}. This mode of encoding may also allow for speed invariant perception of texture observed in humans~\cite{c13}.

Inspired from this, in this work, we compare two approaches evaluated by their performance on naturalistic texture classification: 1) using sensory information about the texture directly encoded by a single channel neuromorphic sensor (mechanoreceptor) and 2) using statistics computed from gray level co-occurrences matrices (GLCM) that encoded the spatio-temporal activation patterns of 16 neuromorphic tactile sensing elements (mechanoreceptors) arranged in a rectangular grid pattern. We find that texture classification accuracy improved significantly with the second spatio-temporal approach. We further investigated if superior performance was due to exploitation of information encoded in the spatio-temporal activation patterns of sensors. For this, we either removed the precise temporal information or systematically disrupted the spatial arrangement of sensors to alter the spatial structure of three-dimensional response volume of the sensor array. We found a significant performance drop in both cases. In addition, we found that the performance with second approach was more robust to variation in sliding speeds at the test time. 

\subsection{Related Work}
In this work, we focus on naturalistic texture classification, texture being an important surface property estimated through touch. Several studies in the past have addressed this problem using different underlying tactile sensor technologies: strain gauges~\cite{c17,c18}, microphones~\cite{c19}, MEMS sensors~\cite{c20, c21}, accelerometers~\cite{c22, c23}, multi-modal sensors~\cite{c24}, capacitive\cite{c25} and optical sensors~\cite{c26, c27}. However, only a handful of studies, have used a neuromorphic spiking code as observed in glabrous fingerpad mechanoreceptors - to encode artificial tactile sensor outputs. 

One such study was done by \textit{Rongala et. al.}~\cite{c8}. They demonstrated the use of neuromorphic sensors for natural texture recognition. They modeled each sensor of their piezoresistive sensor array as an Izhikevich neuron to encode sensor outputs as spike trains. The experimented with two different feature spaces: 1) inter-spike statistics and firing rate to represent individual spike trains and 2) Victor-Purpora distance (VPd) computed between spike trains. These feature spaces combined with KNN algorithm were used for classification. They reported superior performance with inclusion of spike-timing information through VPd. The classification performance reported in this work, however, was based on single channel responses only.

\textit{Rasouli et.al.}~\cite{c10} used a piezoresistive fabric based sensor array to classify ten graded artificial textures. They also modeled the sensors as Izhikevich neurons to mimic mechanoreceptor responses (slow adapting (SA) and fast adapting (FA)) to sensor outputs. The sensor outputs were then fed to a extreme learning machine (ELM) based classifier that was implemented on a custom ELM chip.

\textit{Friedl et. al.}~\cite{c28} implemented a recurrent spiking neural network for classifying 18 metal textures. The neural network comprised of three layers. In first layer, sensor outputs (three accelerometers, two 1-axis and one 3-axis) were converted into spike trains using a heterogeneous population of neurons, tuned to mimic the spiking activity of mechanoreceptor cells. Following this, in second layer, the resultant spiking activity was convolved with bandpass filters to extract nonlinear frequency information from spike trains. Finally, the third recurrent layer implemented a support vector machine (SVM) that was used to classify input into texture classes based on the high-dimensional information obtained from previous layer. 

\textit{Gupta et.al.}~\cite{c9} used a piezoresistive fabric based neuromorphic desktop array~\cite{c10} to classify five graded artificial textures. A high tactile texture image (TTI) was first constructed by superimposing successive texture snapshots encoded in spike events (generated as texture was palpated over the sensor) after compensating for motion parameters. Following this, TTI was transformed into frequency domain using 2-D discrete fourier transform and rotation-invariant features were extracted. These features along with SVM algorithm were used to build a classifier. They found that rotation-invariant features performed better under change in palpation direction at test time.

\textit{Ward-Cherrier et.al.}~\cite{c29} used a neuromorphic variant of the biomimetic optical sensor described in ~\cite{c26} - to classify artificial and natural textures. They compared different spike coding strategies: spatial, temporal and spatio-temporal based on their performance on texture classification. They found that temporal coding (temporal only or spatio-temporal) offered best performance across both artificial and natural textures. 

\section{Material and Methods} 

\subsection{Tactile Sensor}

\begin{figure}[h]
  \centering
  \includegraphics[width=0.45\textwidth]{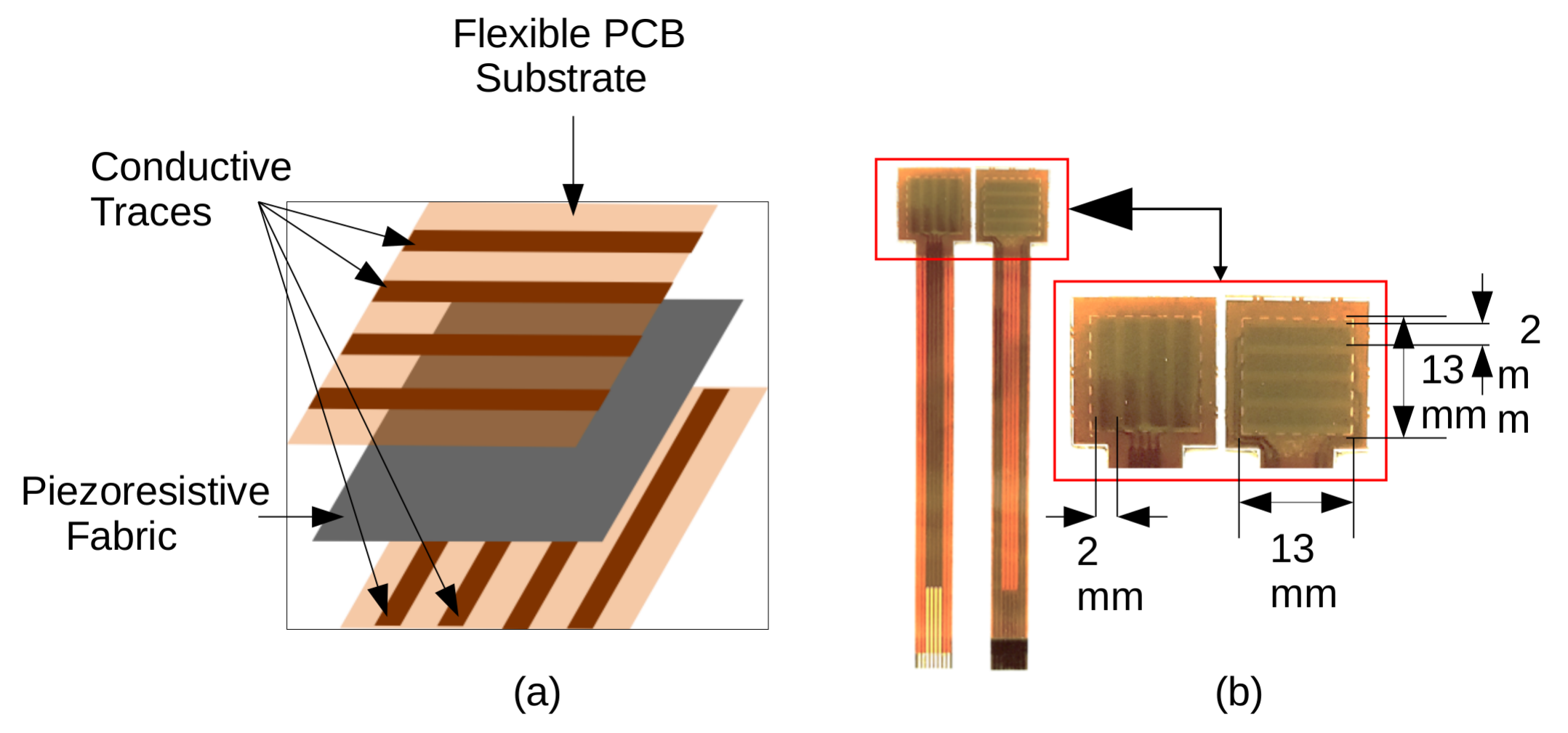}
  \caption{\textbf{Neuromorphic tactile sensor:} a) sensor fabrication principle, b) sensor technical specifications. The sensor was fabricated by sandwiching a piezoresistive fabric (Eeonyx, USA) between conductive traces that were printed on flexible PCB substrate. The sensor has 16 sensing elements (taxels) distributed uniformly in an area of 169 mm$^{2}$. Each individual taxel spanned an area of 4 mm$^{2}$.}
  \label{fig:Sensor}
\end{figure}

A tactile sensor array, similar to one used by \textit{Kumar et.al.} in~\cite{c30}, was fabricated by sandwiching a piezoresistive fabric (LR-SLPA, Eeonyx, CA, USA) between a set of conductive electrodes (see figure~\ref{fig:Sensor}(a)). The electrodes were printed on flexible printed circuit board substrate and were arranged in an orthogonal fashion. Each intersection point between the electrodes and the sandwiched piezoresistive fabric formed a sensing element (taxel). The resultant sensor array had 16 taxels arranged in a rectangular grid and spread over an active sensing area of 169 mm$^2$, with each individual taxel (square) spanning an area of 4 mm$^2$ (see figure~\ref{fig:Sensor}(b)).    

Each taxel was individually sampled at 1000 Hz using a custom built data acquisition board (DAQ). The data was transferred to PC via USB communication~\cite{c30}. All further analysis, that includes conversion of analog data to spike trains (via Izhikevich neuron model), feature extraction and classification (KNN) were done offline in MATLAB on a 2014 MacBook Pro. 

\subsection{Experimental Set-Up}
\begin{figure}[h]
  \centering
  \includegraphics[width=0.48\textwidth]{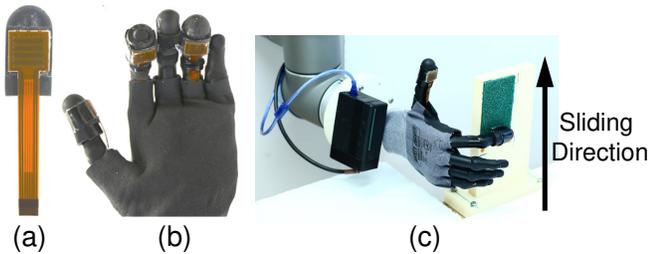}
  \caption{\textbf{Experimental set up:} a) sensor mounted on a soft finger cuff, b) sensor mounted on the end-effector, c) sensorized end-effector mounted on robotic arm (UR10, Universal Robotics, USA). The black box contains the sampling board.}
  \label{fig:SetUp}
\end{figure}

The sensor was first mounted on an anthropomorphic end-effector (iLimb, Touch Bionics, UK) via a 3D printed soft finger cuff (Tango Black, Connex3 Objet260, Stratys, USA, see figures ~\ref{fig:SetUp} (a) \& (b)). In order to protect the sensor as well as to improve contact and diffuse the applied force, the sensor was covered with a transparent soft layer (VHB tape, 3M). Finally, the end-effector along with DAQ was mounted on the UR-10 robotic arm (Universal Robots, USA, see figure ~\ref{fig:SetUp} (c)).

The stimuli were mounted on a 3D printed platform (ABS, Connex3 Objet260, Stratys, USA) by using double-sided tape. The platform, itself, was screwed to the wooden table to restrict motion during the experiment (see figure~\ref{fig:SetUp}(c)). 

For experiments, the sensor was first brought in contact with the stimuli (see figure~\ref{fig:NaturalTextures}). It was then pressed further against the stimuli till the normal forced reached 1 N and kept constant via closed-loop force control. The sensor was then palpated across the stimuli at a constant speed for a distance of 90 mm. The data, here, was gathered for three different palpation speeds: 5, 10 and 15 mm/s. The normal contact force, palpation speeds and distance were kept constant across all the experimental runs.

\subsection{Tactile Stimuli}
\begin{figure}[h]
  \centering
  \includegraphics[width=0.4\textwidth]{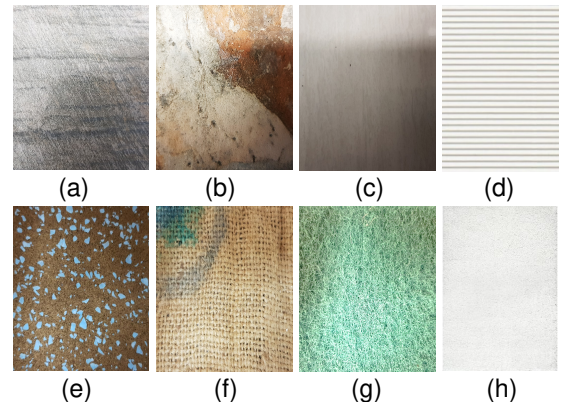}
  \caption{\textbf{Tactile stimuli used for the experiment:} a) floor tile 1, (b) floor tile 2  and (c) floor tile 3, are ceramic floor tiles with different surface profiles, (d) corrugated paper, (e) floor rubber tile, (f) textile rug, (g) scotch brite and (h) styrofoam.}
  \label{fig:NaturalTextures}
\end{figure}

We selected eight different naturalistic textures for this study that included textile, floor tiles and foam (see figure~\ref{fig:NaturalTextures}). The textures were chosen so that there varied widely in their physical properties like friction, roughness and spatial regularity.

\subsection{Experimental Protocol}
\begin{figure}[h]
  \centering
  \includegraphics[width=0.52\textwidth]{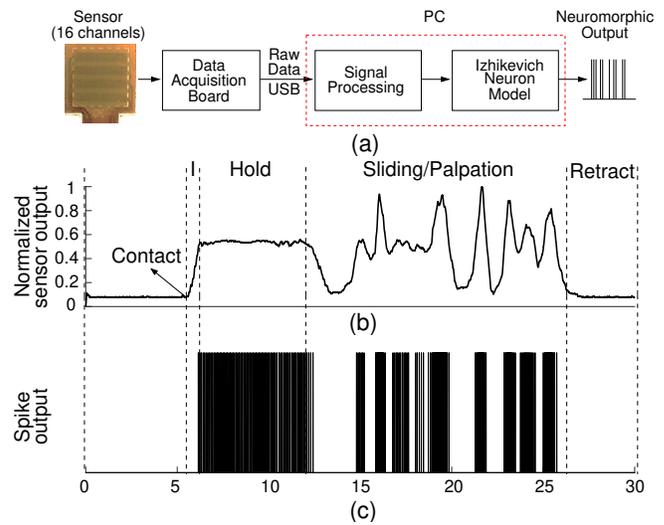}
    \caption{\textbf{Sensor data processing and sensor response in different phases of experimental protocol:} a) schematic of sensor data processing pipeline b) normalized analog sensor output of a representative sensing element at a contact force of 1 N and sliding velocity of 5 mm/s. I here stands for indentation phase, c) spiking activity of artificial mechanoreceptor when injected with signal in b), amplified by gain (G), as input current.}
  \label{fig:protocol}
\end{figure}

The developed neuromorphic system (see figure~\ref{fig:protocol} (a)) was evaluated on its ability to recognize natural textures under a dynamic touch protocol i.e. the stimuli were kept fixed while the sensor was slid across the stimuli to elicit a relative motion at the interface between the two. The experiment comprised of four distinct phases: 1) \textit{Contact:} sensor brought in contact with the stimuli and pressed against the stimuli until a contact force of 1 N was reached, 2) \textit{Hold:} relative motion between the sensor and stimuli was frozen for a period of 5 s to allow sensor to stabilize, 3) \textit{Sliding:} sensor was slid across the stimuli at a constant speed for 90 mm (speeds tested: 5, 10 and 15 mm/s) and 4) \textit{Retract:} break contact between sensor and stimuli and move sensor back to starting position. The experimental protocol is illustrated in figure  ~\ref{fig:protocol} (b) \& (c)). 

In order to investigate the impact on performance of the neuromorphic sensory system on texture recognition under varying sensing conditions, the sliding velocity was varied between 5-15 mm/s at a step size of 5 mm/s. This also allowed us to test the robustness of the proposed method under varying sensing conditions. For each texture and sliding velocity, 20 trials were conducted. The performance on texture recognition was evaluated using a KNN algorithm via 5-fold cross-validation using Euclidean distance.


\subsection{Neuronal Model}
The analog sensor outputs were first low pass filtered to reduce noise. Following this, they were normalized by dividing with the maximum sensor response across all sensing elements. Instead of normalizing each sensing element in isolation, this approach was adopted to ensure that the relative activation levels of the sensing elements remain preserved. 

\begin{figure}[h]
  \centering
  \includegraphics[width=0.48\textwidth]{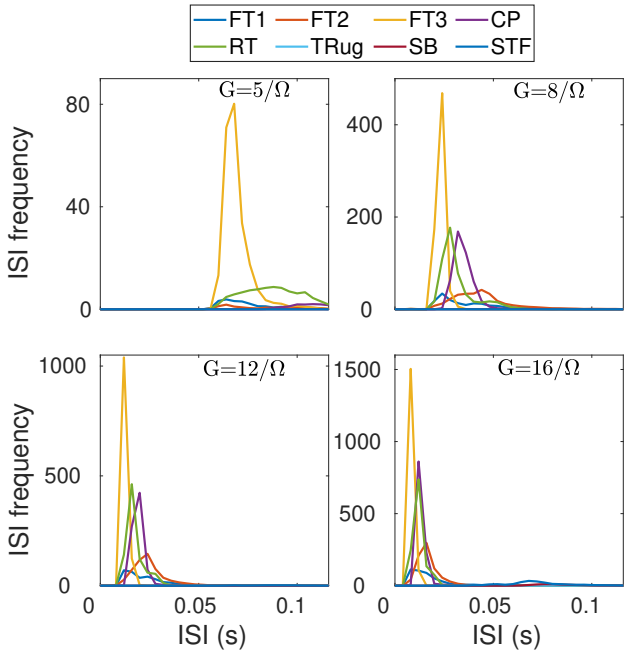}
    \caption{\textbf{ISI distribution as a factor of gain G:} Gain G converts the analog sensor outputs to input current that is fed to Izhikevich neuron model. The plot shows the data collected over the sliding phase of the experimental protocol for all eight stimuli, averaged over the 20 repetitions for each stimuli. Sliding velocity is 5 mm/s. FT1: Floor Tile 1, FT2: Floor Tile 2, FT3: Floor Tile 3, RT: Rubber Tile, TRug: Textile Rug, SB: Scotch Brite, CP: Corrugated Paper, STF: Styrofoam (see figure~\ref{fig:NaturalTextures})}
  \label{fig:isi_GF}
\end{figure}

To convert sensor outputs to spike trains, the pre-processed sensor outputs (SO) were multiplied by a gain (G) and fed as input current ($I_{in}$) to an Izhikevich neuron model~\cite{c31} to mimic the behavior of slow-adapting (SA-I) mechanoreceptors. Similar to prior studies~\cite{c8,c10}, to arrive at optimum value of gain, we computed the inter-spike interval (ISI) histograms of sensor responses with gain values ranging from 5 to 20/$\Omega$. The optimum value of gain was chosen to strike a balance between: 1) increase in firing rate (reduced ISI) with increase in gain resulting in reduced discriminating ability across responses to different stimuli (due to failure in encoding stimuli specific modulations) and 2) sparse responses due to decrease in gain resulting in loss of encoding ability (decrease in representative power) and increase in response latency (see figure~\ref{fig:isi_GF}). Therefore, we selected a gain of 8/$\Omega$ for the experiments in this paper.  

The parameters of the Izhikevich neuron model: membrane potential \textit{v} and adaptation variable \textit{u}, used here to convert analog sensory outputs to spike trains, were updated using differential equations:

\begin{flalign}
\dot{v} &= 0.04v^{2} + 5v + 140 -u + I_{in} \\
\dot{u} &= a(bv-u)
\end{flalign}

\noindent where $I_{in} = \textrm{SO x G}$ is the input current, SO represents analog sensor output voltage post pre-processing and G is the gain. The value of G used here is 8/$\Omega$. A spike is produced when membrane potential reaches 30 mV. This is followed by an after-spike resetting governed by equation:

\begin{flalign}
 \text{if } v \geq 30  \text{mV}, \text{ then} \begin{cases}
  v \leftarrow c \\
  u \leftarrow u + d    
\end{cases}
\end{flalign}

\noindent The parameters in eqs. \textit{2 \& 3} above: a, b, c and d were selected to mimic regular spiking behaviour in neurons with spike-frequency adaptation~\cite{c31}. The values used for these parameters were: $a = 0.02$, $b = 0.2$, $c=-65$ and $d=8$ respectively.

\subsection{Feature Extraction and Classification}

\begin{figure}[h]
  \centering
  \includegraphics[width=0.48\textwidth]{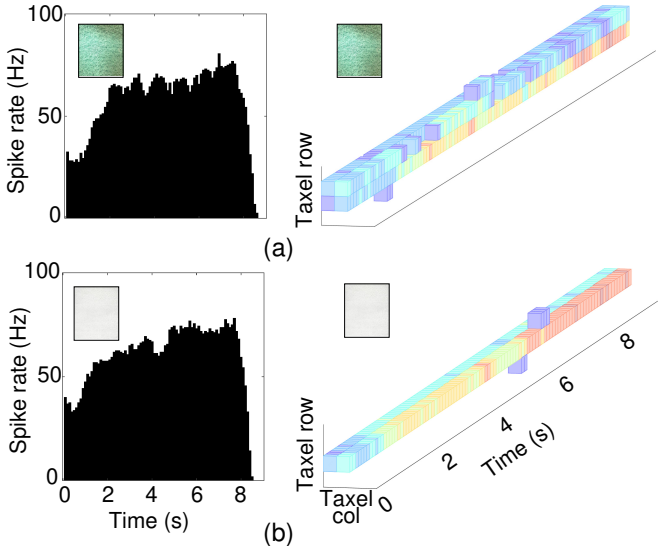}
  \caption{\textbf{Artificial mechanoreceptor response to tactile stimuli:} a) scotch brite and b) styrofoam. The respective textures are shown in inset. The results are for a representative sensing element. Left: post-stimulus time histogram (PSTH), Right: three-dimensional sensor response volume. Each voxel represents response of a single mechanoreceptor at a time slice of 200 ms duration. The color depicts the firing intensity here quantified by spike count. Sliding velocity is 5 mm/s. The side-by-side comparison reveals that 3D voxel responses for two textures are more distinct than their respective PSTH.}
  \label{fig:psth_3dvol}
\end{figure}

As discussed previously, in this study, we compare two approaches \textit{vis-\`{a} vis} their performance on naturalistic (spatially non-uniform) texture classification: 1) \textit{Approach 1 (single taxel):} information encoded by single sensing element spike trains alone is used (see figure~\ref{fig:psth_3dvol}) and 2) \textit{Approach 2 (3D-GLCM):} information embedded in the relative activation patterns of multiple sensing elements (16 and arranged in a 4x4 grid) is used. For both approaches, only the sensor responses from the sliding phase of the experimental protocol were used.

The features used for both approaches were different, however same number of features were used across both approaches for fair performance comparison. For \textit{single taxel} approach, we used three features to represent the sensor responses pertaining to different stimuli: 1) mean spiking: $\mathrm{MSR = n_{spc}/T}$, where $n_{spc}$ is spike count for spike train of length T, 2) coefficient of variation of inter-spike intervals (ISI): $\mathrm{CV_{ISI} = \sigma_{isi}/\mu_{isi}}$, where $\mu_{isi}$ and $\sigma_{isi}$ are mean and standard deviation of ISI distribution and 3) Fano factor: $\mathrm{F = Var_{spc}[W]/n_{spc}[W]}$, to measure longer time scale variability in spike trains. $Var_{spc}$ and $n_{spc}$ are variance and mean spike count for time window of length W.

For the \textit{3D-GLCM} approach, features were not extracted directly from raw spike data but from co-occurrence matrices (GLCM) computed from three-dimensional sensor response volume (see figure~\ref{fig:psth_3dvol}): first two planar dimensions represented the 16 taxels arranged in a rectangular grid (4x4) and the depth dimension represented time with each voxel depth equal to 200 ms. The intensity of each voxel represented the MSR calculated for the appropriate time interval and the sensory channel (taxel).

\subsubsection*{Gray Level Co-occurrence Matrix (GLCM)}
Texture is a global pattern that arises from the repetition of local sub-patterns or features. For this reason, GLCMs~\cite{c32} - that can encapsulate the textures by extracting local patterns/features along with their frequency of occurrence have been extensively used for two- and three-dimensional visual texture classification~\cite{c32,c33}. GLCMs are essentially two-dimensional histograms of the frequency of occurrence of intensity value pairs (local sub-patterns/features) in a given spatial relationship (at a given offset distance and direction, eq.~\ref{eq:glcm}) as a function of intensity level. 

Multiple GLCMs can be constructed by using multiple offset distances and directions to capture local features at multiple scales. A mean gray level co-occurrence matrix (MGLCM) can then be computed by averaging all GLCMs - one for each offset distance and direction (see table~\ref{table:offsetvecs}). This MGLCM, due to averaging across directions, is also rotation invariant. For classification, multiple statistical features (Haralick features) are computed from MGLCM that summarize succinctly the distribution of local features. The GLCM matrix, G is given by:

\begin{equation}\label{eq:glcm}
    G (i, j) = \sum_{x}\sum_{y}\sum_{z} M
\end{equation}

\noindent where, 
$$ M = 
\begin{cases}
    1, & \textrm{if V(x, y, z) = i} \mbox{ \& } \textrm{V(x+$D_x$, y+$D_y$, z+$D_z$) = j} \\
    0, & \text{otherwise}. 
\end{cases}$$

\noindent where $i$, $j$ are the intensity levels; $x, y, z$ are the spatial location in the volume $V$; $D_{x}$, $D_{y}$, $D_{z}$ are offset distances for which co-occurrence matrix ($G$) is computed; $V(x, y, z)$ is value at $(x, y, z)$. $G$ is a $NxN$ GLCM matrix where $N$ is the number of intensity levels.

For current study, four offset distances: 1, 2, 4 and 8 and 13 directions (see table~\ref{table:offsetvecs}) were used to compute three-dimensional GLCMs. These fifty two GLCMs were averaged and then normalized to compute MGLCM. Three Haralick features~\cite{c32} were then computed from mean co-occurrence matrix as features: 

\begin{equation}
    \label{eq:con}
    \textrm{Contrast, C} = \sum_{i} \sum_{j} p(i, j) (i-j)^2
\end{equation}

\begin{equation}
    \label{eq:corr}
    \textrm{Correlation, c} = \sum_{i} \sum_{j} p(i, j) \dfrac {(i-\mu_{x})(j - \mu_{y})}{(\sigma_{x})(\sigma_{y})}
\end{equation}

\begin{equation}
    \label{eq:asm}
     \textrm{Angular second moment, ASM} = \sum_{i} \sum_{j} p(i, j)^2
\end{equation}

\noindent where $p(i, j)$ is the $(i, j)^{th}$ element of the normalized GLCM, $p(x, y)$; $\mu_{x}$, $\mu_{y}$, $\sigma_{x}$, $\sigma_{y}$ are the means and the standard deviations of marginal probability distributions $p_{x}$ and $p_{y}$, computed using equations below:

\begin{equation}\label{eq:mean_std}
    \begin{aligned}
        \mu_{x(y)} &= \sum_{i} \sum_{j} i(j)(p(i, j)),\hspace{1mm} \sigma_{x(y)}^2 \\ &= \sum_{i} \sum_{j}p(i, j)(i(j) - \mu_{x(y)})^2
    \end{aligned}
\end{equation}

Finally, for classification, we used k-nearest neighbours algorithm (KNN) with k=5 neighbours, five-fold cross validation and euclidean distance. 

\begin{table}[h]
\centering
    \caption{Offset vectors to generate co-occurrence matrices. Offset distance, D values  used were 1, 2, 4 $\&$ 8.}
    \label{table:offsetvecs}
    \begin{tabular}{|c|c|c|c|c|c|} \hline
    \begin{tabular}[c]{@{}c@{}}ID\\ \end{tabular} & \begin{tabular}[c]{@{}c@{}}Offset \\ Vector\end{tabular} & \begin{tabular}[c]{@{}c@{}}Direction\\($\theta, \phi$)\end{tabular} & ID & \begin{tabular}[c]{@{}c@{}}Offset\\Vector\end{tabular} & \begin{tabular}[c]{@{}c@{}}Direction\\($\theta$, $\phi$)\end{tabular} \\ \hline
    1 & (0, D, 0) & (0$\degree$, 0$\degree$) & 8 & (-D, 0, -D) & (90$\degree$, 45$\degree$) \\ \hline
    2 & (-D, D, 0) & (45$\degree$, 0$\degree$) & 9 & (D, 0, -D) & (90$\degree$, 135$\degree$) \\ \hline
    3 & (-D, 0, 0) & (90$\degree$, 0$\degree$) & 10 & (-D, D, -D) & (45$\degree$, 45$\degree$) \\ \hline
    4 & (-D, -D, 0) & (135$\degree$, 0$\degree$) & 11 & (D, -D, -D) & (45$\degree$, 135$\degree$) \\ \hline
    5 & (0, D, -D) & (0$\degree$, 45$\degree$) & 12 & (-D, -D, -D) & (135$\degree$, 45$\degree$) \\ \hline
    6 & (0, 0, -D) & (0$\degree$, 90$\degree$) & 13 & (D, D, -D) & (135$\degree$, 135$\degree$) \\ \hline
    7 & (0, -D, -D) & (0$\degree$, 135$\degree$) &  &  &  \\ \hline
    \end{tabular}
\end{table}

\section{Results}

\subsection{Texture Classification}
As discussed before, we compared the relative performance of two approaches on tactile texture classification task : 1) \textit{Approach 1 (single taxel)} used texture information encoded by a single sensing element, represented using mean spiking rate (MSR), coefficient of variation of inter-spike intervals and fano factor. Apart from MSR, other two statistics were used to capture temporal information encoded by the sensing element and 2)  \textit{Approach 2 (3D-GLCM}) used statistics like contrast, correlation and angular second momentum (ASM) extracted from three-dimensional co-occurrence matrices (3D-GLCM) that encoded the texture information embedded in the spatio-temporal mechanoreceptor activation patterns. This approach allowed to capitalize fully on the spatial and temporal information encoded by artificial mechanoreceptors (sensing elements) for classification.

\begin{figure}[h]
  \centering
  \includegraphics[width=0.48\textwidth]{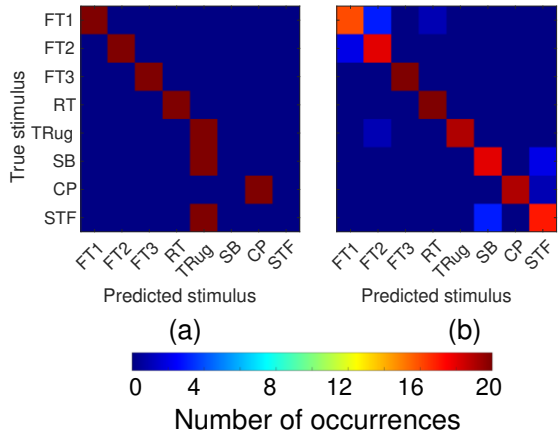}
  \caption{\textbf{Classification Errors:} a) confusion matrix for single taxel approach, b) confusion matrix for 3D-GLCM approach. As evident, the 3D-GLCM approach is able to better disambiguate texture identity (compare predictions for SB, CP, STF) than the single taxel approach - resulting in higher performance on texture classification task. Sliding speed was 5 mm/s. FT1: Floor Tile 1, FT2: Floor Tile 2, FT3: Floor Tile 3, RT: Rubber Tile, TRug: Textile Rug, SB: Scotch Brite, CP: Corrugated Paper, STF: Styrofoam (see figure~\ref{fig:NaturalTextures})}
  \label{fig:confmat}
\end{figure}

We found that \textit{3D-GLCM} approach consistently outperformed \textit{single taxel} approach across multiple sliding velocities (see table~\ref{table:classifyacc}). The margin of the outperformance was as high as 22.7\% at 5 mm/s sliding velocity, dropping to 17\% at 15 mm/s. This outperformance was primarily due to better classification performance of  \textit{3D-GLCM} approach on three textures namely: rug, scotch brite and styrofoam (see figure~\ref{fig:NaturalTextures}). The \textit{single taxel} approach failed to successfully resolve these textures and misclassified scotch brite and styrofoam as rug (see figure~\ref{fig:confmat}. We also found that the outperformance was robust to the number of neighbours (K) used in KNN for classification.  
\begin{table}[h]
    \centering
    \caption{Classification Performance Comparison between single taxel \& 3D-GLCM approaches.}
    \label{table:classifyacc}
    \resizebox{1\linewidth}{!}{%
    \begin{tabular}{|c|c|c|c|c|} 
    \hline
     & \multicolumn{4}{c|}{\begin{tabular}[c]{@{}c@{}}Classification Accuracy (\%)\end{tabular}} \\ 
    \hline
    \multirow{4}{*}{\begin{tabular}[c]{@{}c@{}}Sliding~\\Velocity\\(mm/s)\\ \end{tabular}} &  & Single Taxel & 3D-GLCM & \begin{tabular}[c]{@{}c@{}}Change (\%)\end{tabular} \\ 
    \cline{2-5}
     & 5 & 75 & 92 & 22.7 \\ 
    \cline{2-5}
     & 10 & 73 & 88 & 20.5 \\ 
    \cline{2-5}
     & 15 & 72 & 84 & 17.0 \\
    \hline
    \end{tabular}
    }
\end{table}

\subsection{Spatial Location Perturbation}

It was hypothesised that capturing of spatio-temporal patterns in firing activity of artificial mechanoreceptors was primarily behind the superior performances of \textit{3D-GLCM} approach. 

\begin{figure}[h]
  \centering
  \includegraphics[width=0.45\textwidth]{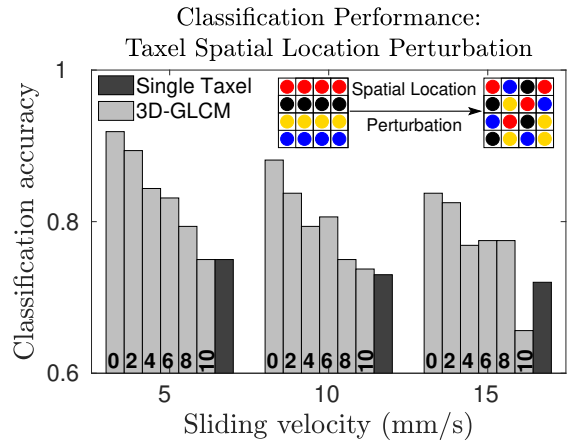}
  \caption{\textbf{Effect of artificial mechanoreceptor spatial location perturbation on classification performance:} The light grey bars show the classification accuracy obtained using 3D-GLCM approach when spatial location of mechanoreceptors was perturbed. The value on the bottom of the bars show the number of pixels that were perturbed. The dark grey bar shows the classification accuracy obtained using single taxel approach. The performance of 3D-GLCM approach drops as level of perturbation increases, depicting the importance of spatial structure of sensor response on classification performance.}
  \label{fig:pp}
\end{figure}

In order to evaluate the importance of preserving the spatial relationships between artificial mechanoreceptors on the classification performance, we systematically perturbed their spatial location. The temporal activity of any individual mechanoreceptor, however, was left unperturbed. To spatially perturb the mechanoreceptors, we first selected \textit{n} number of mechanoreceptors and then randomly reassigned new locations to them in the 4x4 sensor grid (see figure~\ref{fig:pp}). This alteration of the spatial structure of the sensor response resulted in a significant performance degradation.


As the order of perturbation increased (more mechanoreceptors perturbed), the classification performance dropped either to levels of performance exhibited by \textit{single taxel} approach or even below (see figure~\ref{fig:pp}), possibly due to the injection of spurious relationships by way of spatial perturbations in the sensory response. This clearly demonstrates the importance of response\textquotesingle s spatial structure on classification performance. 

\subsection{Temporal Information}
As a complimentary analysis to the one in previous section, here we evaluated the importance of encoded temporal information about the stimuli towards classification performance. For this, we collapsed the time dimension of the three-dimensional sensor response volume (see figure~\ref{fig:psth_3dvol}), removing the information encoded in temporal firing patterns. Following this, each mechanoreceptor response was represented by mean spiking rate of the full signal. We then computed the co-occurrence matrices from the two-dimensional representation of the sensor response (2D-GLCM) and extracted same features as 3D-GLCM to compare classification performance.  

As expected, the 2D-GLCM approach fared worse than the
\textit{3D-GLCM} approach as it ignores the precise temporal information encoded in temporal firing patterns (see table~\ref{table:tempinfo}) as described above. However, the \textit{2D-GLCM} approach performance fell below the \textit{single taxel} approach as well (see table~\ref{table:tempinfo}). One possible cause for this could be that the distinctive information, required to classify the textures used in this study, is higher in the temporal dimension than the spatial dimension. Thus, although, the temporal information was only fully utilised in 3D-GLCM, the use of coefficient of variation of inter-spike intervals and fano factor as features (these features encode temporal information), resulted in better performance of the \textit{single taxel} approach \textit{vis-\'{a}-vis} 2D-GLCM.




\begin{table}[h]
    \centering
    \caption{Impact on classification performance on removal of temporal information.}
    \label{table:tempinfo}
    \resizebox{\linewidth}{!}{%
    \begin{tabular}{|c|c|c|c|c|} 
    \hline
     & \multicolumn{4}{c|}{\begin{tabular}[c]{@{}c@{}}Classification Accuracy\\(\%)\end{tabular}} \\ 
    \hline
    \multirow{4}{*}{\begin{tabular}[c]{@{}c@{}}Sliding~\\Velocity\\(mm/s)\\ \end{tabular}} &  & Single Taxel & 2D-GLCM & 3D-GLCM \\ 
    \cline{2-5}
     & 5 & 75 & 46 & 92 \\ 
    \cline{2-5}
     & 10 & 73 & 33 & 88 \\ 
    \cline{2-5}
     & 15 & 72 & 43 & 84 \\
    \hline
    \end{tabular}
    }
\end{table}

\subsection{Time To Recognition (TOR)}

As important as it is to achieve high accuracy of classification, it is also important to achieve the same in minimal time (low latency/TOR). The time here refers to the time required to interact with the stimuli to encode enough information to carry out successful classification and not the computation time. Low latency is, in particular, very important for real-time applications. 

We, thus, compared the two approaches on the time required for achieving a given classification accuracy: here chosen to be the accuracy obtained with the \textit{single taxel} approach when full signal is used for classification. We hypothesised that, \textit{3D-GLCM} approach, on account of using richer information will achiever similar classification performance as the \textit{single taxel} approach in only a fraction of the sliding time.  

\begin{figure}[h]
  \centering
  \includegraphics[width=0.45\textwidth]{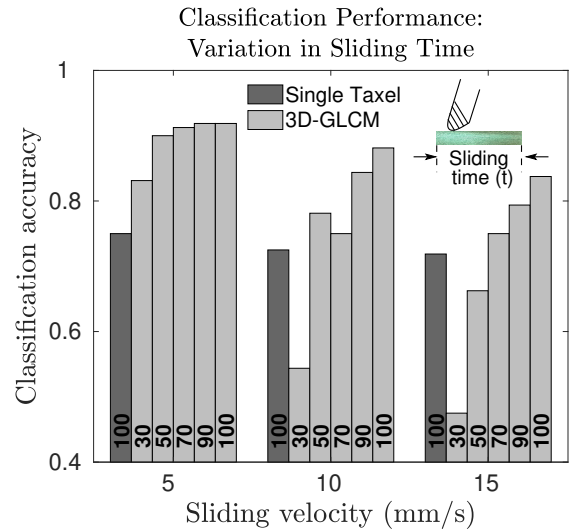}
  \caption{\textbf{Time To Recognition:} The dark grey bar shows the classification accuracy obtained using single taxel approach with full signal. In comparison, the light gray bars show the performance obtained using 3D-GLCM with varying duration of sliding time. The values on the bottom of the bars represent the percentage of the total sliding time used for classification. As evident from the plot, the 3D-GLCM approach is able to match the classification performance of single taxel approach in a fraction of the sliding time, t.}
  \label{fig:ttr}
\end{figure}

As hypothesised, we found that \textit{3D-GLCM} approach indeed took a fraction of time (sliding time) taken by the \textit{single taxel} approach to match its classification performance (see figure~\ref{fig:ttr}). This phenomenon of outperformance in the time taken to classify was robust across multiple sliding velocities tested. However, the scale of outperformance was dependent on the sliding velocity with less time taken at lower velocity (30\% at 5 mm/s) than higher sliding velocity (70\% at 15 mm/s) to match the performance of \textit{single taxel} approach by the \textit{3D-GLCM} approach.

\subsection{Invariance to Sliding Velocity}
In real-life deployment scenarios, the sensing conditions are not known a priori. It is therefore essential for a system, here texture classification system, to adapt and retain performance when sensing conditions differ from training conditions. Here, we evaluated, our proposed texture classification system, \textit{3D-GLCM} approach, under varying sliding velocities. For this, the system was trained on two sliding velocities and tested on the remaining sliding velocity. To serve as reference, the same procedure was repeated for the \textit{single taxel} approach.

We hypothesised that the spatio-temporal firing patterns of mechanoreceptors should remain unchanged across varying sliding velocities, although firing patterns of individual mechanoreceptors may change, hence \textit{3D-GLCM} approach should perform better than the \textit{single taxel} approach. We, as hypothesized, found that the \textit{3D-GLCM} approach indeed outperformed the \textit{single taxel} approach (see table~\ref{table:speedvar}). 

\begin{table}
    \centering
    \caption{Invariance to Sliding velocity}
    \label{table:speedvar}
    \resizebox{\linewidth}{!}{%
    \begin{tabular}{|c|c|c|c|c|} 
    \hline
     & \multicolumn{4}{c|}{\begin{tabular}[c]{@{}c@{}}Classification Accuracy\\(\%)\end{tabular}} \\ 
    \hline
    \multirow{4}{*}{\begin{tabular}[c]{@{}c@{}}Test\\Sliding~\\Velocity\\(mm/s)\\ \end{tabular}} &  & Single Taxel & 3D-GLCM & \multicolumn{1}{c|}{\begin{tabular}[c]{@{}c@{}}Change\\(\%)\end{tabular}} \\ 
    \cline{2-5}
     & 5 & 28 & 54 & 93 \\ 
    \cline{2-5}
     & 10 & 30 & 51 & 70 \\ 
    \cline{2-5}
     & 15 & 27 & 40 & 48 \\
    \hline
    \end{tabular}
    }
\end{table}

\section{Discussion}
In this work, we proposed a neuromorphic tactile system for naturalistic texture classification that used a spatio-temporal approach to represent artificial mechanoreceptor responses to tactile  stimuli. 

This spatio-temporal approach (3D-GLCM) was executed by computing co-occurrence matrices from three-dimensional sensor response volume -- each voxel represented by mean spike rate of 200 ms of sensor response. The performance of the \textit{3D-GLCM} approach was compared to two approaches: 1) purely spatial approach (2D-GLCM) also based on computation of co-occurrence matrix from sensor response -- each cell represented by mean spike rate of full signal and 2) approach based on just the response of single sensing element (single taxel). 

Firstly, we found that \textit{3D-GLCM} approach exhibited better performance than both \textit{2D-GLCM} and \textit{single-taxel} approaches on naturalistic texture classification task. This outperformance was dependent on relative sliding velocity between the sensor and the textural stimuli with reduction in margin of outperformance with increase in sliding velocity. For example, the margin of outperformance between \textit{3D-GLCM} and \textit{single-taxel} approaches was 22.7\% at 5 mm/s and 17\% at 15 mm/s respectively (see table~\ref{table:classifyacc}).
 
Secondly, we evaluated the impact of each of the sensory dimensions of proposed spatio-temporal approach on classification performance. To investigate the importance of retaining precise spatial relationships between firing activity of mechanoreceptors, we gradually perturbed the spatial location of mechanoreceptors to alter the spatial structure of the sensory response volume (see figure~\ref{fig:psth_3dvol}(b)). This resulted in sharp drop in performance which increased with increase in level of perturbation. Similarly, when temporal information was removed (2D-GLCM), the performance dropped in excess of 40\% across all sliding velocities (see table~\ref{table:tempinfo}).  

Finally, we also found that \textit{3D-GLCM} approach was both more robust to change in sensing conditions at test time (here change in test time sliding velocity, see table~\ref{table:speedvar}) as well as achieved equivalent performance to the \textit{single-taxel} approach within only a fraction of the sliding time (see figure~\ref{fig:ttr}). 

\section{Conclusions and Future Work}
The results presented in this work clearly demonstrate that the use of population responses (spatio-temporal) over individual sensor/mechanoreceptor responses is a superior way of encoding and representing information about the tactile stimuli. The spatio-temporal strategy may not only provide better absolute task performance -- for example here on texture classification, but is also better equipped to adapt to change in sensing/task conditions over time. Although, the spatio-temporal \textit{3D-GLCM} approach proposed in this work performed better than both chance level and \textit{single taxel}, the classification accuracy achieved was low. This clearly indicates a necessity for a more effective spatio-temporal approach. For example, an approach that can successfully disentangle sensor response into its constituents: 1) component representing textural features and 2) component representing the transformations in sensor response due to variation in sliding speed, may offer better performance. In addition, this approach may better adapt to out of distribution samples. This approach may also be extendable to include other stimuli transformations like indentation force, spatial transformations like pose as well as to other tactile tasks. 

The proposed \textit{spatio-temporal strategy} offers better performance than \textit{single-taxel} approach albeit at a possibly higher computational cost. Though we have not focused on the aspect of computational cost associated with different approaches in this work, nonetheless it is an important consideration. There is a trade-off between computational cost and performance. An application that needs higher absolute performance would necessitate the use of \textit{3D-GLCM} approach. However, if lower computational cost is desired even if it comes at the expense of lower performance then \textit{single-taxel} approach would be the ideal choice. An important focus of future work should thus be to develop spatio-temporal strategies that offer higher performance with similar or lower computational costs to \textit{single-taxel} approach.      

In this work the sensing elements were modeled after slow-adapting mechanoreceptors, however, neurophysiological studies have shown that both slow-adapting and fast-adapting receptors are involved in texture recognition in humans~\cite{c34}. Thus, higher performance may be obtained using an approach that can capitalize on and combine information encoded by different types of mechanoreceptors that respond to different frequency bands. For example, extracting underlying harmonic structure in spatially distributed differently frequency tuned mechanoreceptors may offer better invariance to sliding speed. 

An obvious drawback of our approach is its reliance on whole spike trains making it inherently non real-time. In addition, we were not able to develop an end-to-end spatio-temporal neuromorphic system and had to transform neuromorphic data for feature extraction and classification unlike in~\cite{c28}. Such end-to-end neuromorphic system will not only allow implementation of brain-inspired algorithms for tactile tasks but may also reduce the computational requirements/cost of the implementations.

Finally, one drawback on the hardware side was the arrangement of sensors in a uniform rectangular grid that results in acquisition of redundant information. A better strategy would, thus, be to arrange sensors non-uniformly similar to distribution of mechanoreceptors in human fingertip.

In conclusion, although the proposed approach has several drawbacks as discussed above, it shows a clear path forward for spatio-temporal population response based approaches over single sensor response based approaches. It remains to be seen how well current and future spatial-temporal strategies perform on varied tactile tasks.

\section*{ACKNOWLEDGEMENT}
The authors would like to thank Dr. Deepesh Kumar for helpful scientific discussions and Aravindh N. Swaminathan for helping out with experiments. 


\end{document}